\documentclass[runningheads]{llncs}
\usepackage[dvipsnames, svgnames, x11names]{xcolor}
\usepackage{graphicx}
\usepackage{caption}
\usepackage{changes}
\usepackage{multirow}
\usepackage{graphicx}
\usepackage{amsmath,amssymb,bm}
\usepackage{booktabs}
\usepackage{bbding}
\usepackage{subfigure}
\usepackage[pagebackref=true,breaklinks=true,colorlinks,bookmarks=false]{hyperref}

\begin{document}
\title{Black-box Source-free Domain Adaptation via Two-stage Knowledge Distillation 
\thanks{The short version is accepted by IJCAI \href{https://sites.google.com/view/glow-ijcai-23/home}{1st International Workshop on 
Generalizing from Limited Resources in the Open World}}
}
\titlerunning{TKD}

\author{Shuai Wang\inst{1}, Daoan Zhang\inst{2}, Zipei Yan\inst{3}, Shitong Shao\inst{4}, Rui Li\inst{1}}
\institute{Tsinghua University
          \and Southern University of Science and Technology
          \and The Hong Kong Polytechnic University
          \and Southeast University}
\authorrunning{Wang et al.}
\maketitle              % typeset the header of the contribution
\begin{abstract}
    Source-free domain adaptation aims to adapt deep neural networks using only pre-trained source models and target data. However, accessing the source model still has a potential concern about leaking the source data, which reveals the patient's privacy. In this paper, we study the challenging but practical problem: black-box source-free domain adaptation where only the outputs of the source model and target data are available. We propose a simple but effective two-stage knowledge distillation method. In Stage \uppercase\expandafter{\romannumeral1}, we train the target model from scratch with soft pseudo-labels generated by the source model in a knowledge distillation manner. In Stage \uppercase\expandafter{\romannumeral2}, we initialize another model as the new student model to avoid the error accumulation caused by noisy pseudo-labels. We feed the images with weak augmentation to the teacher model to guide the learning of the student model. Our method is simple and flexible, and achieves surprising results on three cross-domain segmentation tasks.
\end{abstract}

\section{Introduction}
\textbf{D}eep \textbf{N}eural \textbf{N}etwork\textbf{s} (DNNs) have achieved remarkable success in the medical image analysis field~\cite{Isensee2020,unet,Esteva2017,Rajpurkar2022,wang2023prototype}. However, DNNs are notoriously sensitive to the domain shift that train and test samples have different distributions. Especially in the medical image analysis field, domain shift usually occurs because medical images are obtained with varying parameters of acquisition or modalities. To tackle this problem, unsupervised domain adaptation (UDA) that aims to transfer knowledge from labeled source domain to unlabeled target domain has been widely explored~\cite{chen2020unsupervised,tsai2018learning,tzeng2017adversarial}.
\begin{table}[t]
    \begin{minipage}{0.4\textwidth}
      \caption{\textbf{Different settings}. UDA: unsupervised domain adaptation. SFDA: source-free domain adaptation. BSFDA: black-box source-free domain adaptation. S. denotes source domain.}
      \begin{center}
      \resizebox{0.9\textwidth}{!}{
        \begin{tabular}{lrr}
               \toprule
              & \multicolumn{1}{l}{S.data} & \multicolumn{1}{l}{S.model} \\
                \midrule
              UDA   & \Checkmark     & \Checkmark \\
              SFDA  &  \XSolidBrush    & \Checkmark \\
              BSFDA & \XSolidBrush      & \XSolidBrush  \\
                \bottomrule
          \end{tabular}
      
      }    
          \label{tab:setting}
    \end{center}
    \end{minipage}
    \hfill
    \begin{minipage}{0.55\textwidth}
      \centering
      \includegraphics[width=\linewidth]{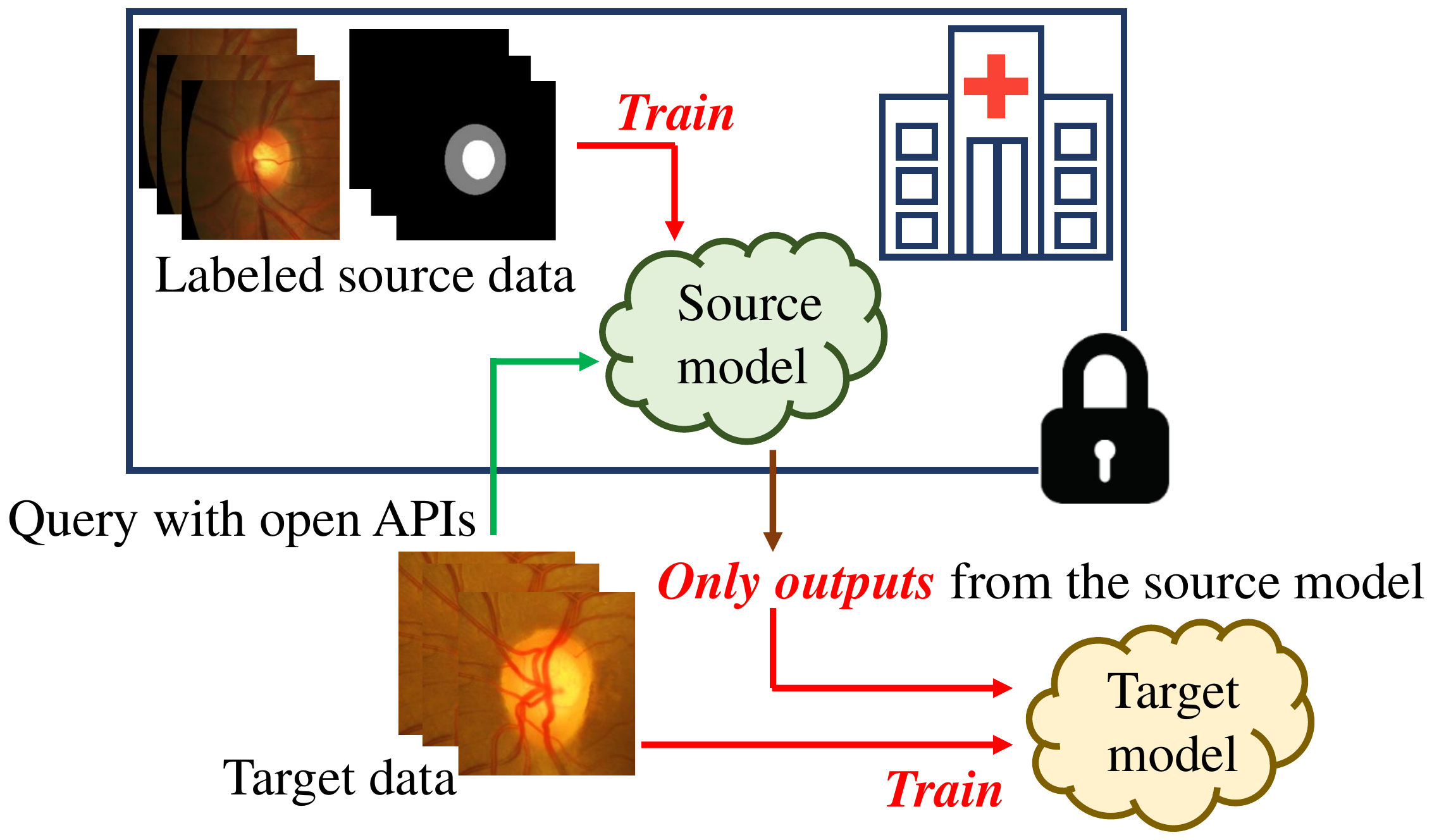}
      %\vfill
      \captionof{figure}{The illustration of challenging but practical problem: black-box source-free domain adaptation setting.
                         }
       \label{fig:setting}
    \end{minipage}
\end{table}

However, existing UDA methods usually need to access the source data, which raises concerns about data security and fairness. Source-free domain adaptation (SFDA) \cite{chen2021source,liang2020we,liu2021source,yang2021generalized,Wang_2023_CVPR} aims to update the model trained on source domain with target data, i.e. not access to the source data. Even so, these methods still require source models, which have two limitations~\cite{liang2022dine}. \textit{First}, it is possible to recover the source data using generative models \cite{gao2022back,gan}. This property may raise potential data security problems. \textit{Second}, SFDA methods usually tune the parameters of the source model. So the target model must employ the same network architecture as the source model, which is impractical for low-source target users, i.e. some community hospitals.

This paper studies the challenging but practical problem: black-box source-free domain adaptation~\cite{liang2022dine} (BSFDA).
The illustration of BSFDA is shown in Figure \ref{fig:setting}. In this setting, we could only access the prediction from the source model via cloud open APIs (e.g. chatGPT \footnote{\href{https://chat.openai.com/}{https://chat.openai.com/}}) and target data without ground truth. Our goal is to train the target model from scratch (or only use ImageNet pre-trained weights \cite{deng2009imagenet}) using target data and the corresponding prediction from the black-box source model. Compared with UDA or SFDA, BSFDA does not raise concerns about data security and model security, as shown in Table \ref{tab:setting}.

To tackle this challenging problem, we propose a novel method based on knowledge distillation \cite{hinton2015distilling}. Our approach consists of two stages. In the first stage, we get soft pseudo-labels from the black-box source model. Then we adopt these soft pseudo-labels to guide the learning of the target model. We train the target model from scratch (or use ImageNet~\cite{deng2009imagenet} pre-traind initialization) with a knowledge distillation~\cite{hinton2015distilling} manner because soft labels usually provide more helpful information (``dark'' knowledge \cite{hinton2015distilling}) than hard labels. The target model trained in the first stage has already achieved promising results but is still suboptimal. This is because pseudo-labels from the source model can not be updated, and the quality of pseudo-labels is never improved. To boost the performance of the target model, we design the two-view knowledge distillation to achieve this. We load the well-trained model in the first stage as the new teacher model and randomly initialize another model as the student model to avoid error accumulation. The key insight in the second stage is that we use soft labels under weak augmentation to guide the training for the strong augmented images. Specifically, we feed the weak augmented images to the teacher network and strong augmented images to the student network. We use the outputs of the teacher model to guide the outputs of the student model because inputs of the teacher network suffer from less distortion. 

We validate our method on three cross-domain medical image segmentation tasks. Our method is simple but achieves state-of-the-art performance under black-box source-free domain adaptation setting on three tasks.
\section{Method}
\begin{figure}[t]
    \includegraphics[width=\textwidth]{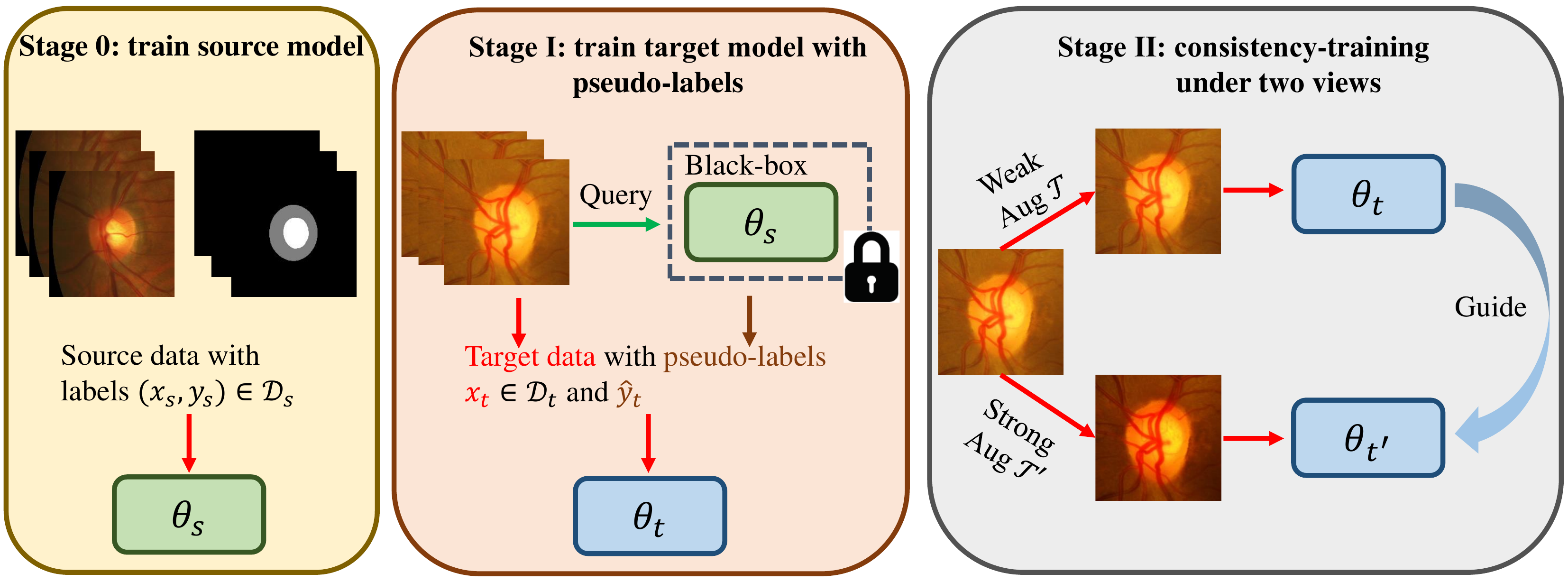}
    \caption{\textbf{The overview of our method}. At Stage 0, we train the source model without bells and whistles. After that, we propose a method called two-stage knowledge distillation to tackle the challenging problem: black-box source-free domain adaptation.}
    \label{fig:framework}
\end{figure}
The overview of our method is shown in Figure~\ref{fig:framework}. Given a set of source images and pixel-wise segmentation labels $(x_s,y_s)\in \mathcal{D}_s$, we first train the source model with parameters $\theta_s$ without bells and whistles. After that, only the trained source model is provided without access to the source data and model parameters. Specifically, only the outputs of the source model for target images $x_t$ are provided to utilize for domain adaption in the target domain. We aim to train a model from scratch (or only use ImageNet pre-trained~\cite{deng2009imagenet} initialization) just using the outputs of the black-box source model and target images $x_t$ with the proposed method, i.e. two stage knowledge distillation. We will describe the details below.
\subsection{Source Model Training}
First, we train the source model on labeled source data $\mathcal{D}_s$ with source images $x_s$ and paired labels $y_s$.
Unlike \cite{bateson2020source,karani2021test}, we train the source model using common cross-entropy loss without bells and whistles.
The objective function is formulated as
\begin{equation}
    \mathcal{L}_s = -\mathbb{E}_{(x_s,y_s)\in \mathcal{D}_s} y_s \log f_{s}(x_s),
    \label{eq:loss_source}
\end{equation}
where $f_s$ denotes the source model with parameter $\theta_s$.

After that, the source model and source data are not accessible. Only the outputs of the source model can be utilized.
\subsection{Two-stage Knowledge Distillation}
\label{sec:two_stage_kd}
For the target domain image $x_t$, we can get the soft pseudo-labels $\hat{y}_t=f_{s}(x_t)$ using the black-box source model $f_{s}$ with open APIs. It is not trivial to design a method to train a new model from scratch only using pseudo-labels provided the source model. A simple strategy to train the target model is self-training that uses pseudo-labels $\hat{y}_t$ with cross-entropy loss. However, two significant concerns will raise. \textit{First}, pseudo-labels are inevitably noise due to the distribution shift between the source domain and the target domain. How to efficiently use knowledge from the source model is still unclear. \textit{Second}, pseudo-labels are frozen because the source model can not be updated after source training (i.e. stage 0).

To address these two issues, we propose a novel method consisting two stages via knowledge distillation. In Stage \uppercase\expandafter{\romannumeral1}, we train the target model $f_t$ from scratch using soft pseudo-labels rather than hard labels which aims to exact more help knowledge from source domain. In Stage \uppercase\expandafter{\romannumeral2}, another model is initialized randomly (or ImageNet~\cite{deng2009imagenet} pre-trained initialization) to avoid error accumulation. After that, we use pseudo-labels under weak data augmentation to guide the learning for the strong augmented images. Next we will describe more details of our method.

\noindent
\textbf{Stage \uppercase\expandafter{\romannumeral1}}.  In this stage, we use knowledge distillation~\cite{hinton2015distilling} to exact knowledge from source model. 
The reason of applying knowledge distillation manner (i.e. soft label) rather than hard pseudo labels have two folds.
\textit{First}, soft label could provide ``dark'' knowledge~\cite{hinton2015distilling} from the source model.
\textit{Second}, soft pseudo-labels work better than hard pseudo-labels for out-of-domain data as suggested in~\cite{noisy_student}.

To be specific, we train the target model $f_t$ with parameter $\theta_t$ from scratch as follows
\begin{equation}
    %\mathcal{L} = -\mathbb{E}_{x_t \in \mathcal{D}_t} \hat{y}_t \log f_{t}(x_t).
    \mathcal{L}_{1} = D_{\textrm{KL}}(\hat{y}_t || f_{t}(x_t)),
    \label{eq:loss_stage1}
\end{equation}
where $D_{\textrm{KL}}$ denotes Kullback-Leibler divergence. This method has certain effect but the model $f_t$ is trained on the target domain with \textit{noisy} and \textit{fixed} labels $\hat{y}_t$, which is suboptimal for target domain. 
Thus, we propose to leverage the second stage to enhance the trained model $f_t$ rely on knowledge distillation between two views of images. 

\noindent
\textbf{Stage \uppercase\expandafter{\romannumeral2}}.  In Stage \uppercase\expandafter{\romannumeral2}, we aim to improve the performance of model $f_t$ via knowledge distillation method between two views of images. We initialize another model $f_{t^{\prime}}$ with parameter $\theta_{t^{\prime}}$ randomly or using ImageNet~\cite{deng2009imagenet} pre-trained weights to avoid error accumulation due to noisy labels in Stage \uppercase\expandafter{\romannumeral1}. The core idea of this stage is to use pseudo-labels of weak augmentation images to guide the learning for the student model. Specifically, let $\mathcal{T}(x_t)$ and $\mathcal{T}^{\prime}(x_t)$ denote the weak and strong augmented images for $x_t$, respectively. We feed the weak augmented images $\mathcal{T}(x_t)$ into $f_t$ to get pseudo-labels $\hat{y}^{\prime}_t = f_t(\mathcal{T}(x_t))$. After that, we use $\hat{y}^{\prime}_t$ to guide the learning of strong augmented images $\mathcal{T}^{\prime}(x_t)$ for model $f_{t^{\prime}}$ because weak augmented images usually produce more reliable pseudo-labels. The loss function of this stage is formulated as follows
\begin{equation}    
    \mathcal{L}_{2} = D_{\textrm{KL}}(\hat{y}^{\prime}_t || f_{t'}(\mathcal{T}^{\prime}(x_t))).
\end{equation}
Finally, we get the target model $f_{t^{\prime}}$ for evaluation.

\noindent
\textbf{Remark}.  Our method has three advantages. \textit{First}, our method does not require the target model to have the same network architecture as the source model, which is helpful for low-source target users (e.g. some community hospitals). \textit{Second}, the final target model $f_{t^{\prime}}$ is \textit{fully} trained on target images $x_t$, which is not affected by domain shift. \textit{Third}, in this setting, parameters of source model $f_s$ are not accessible for target users, which ensures the security of (source) model parameters.

\begin{table}[t]
    \centering
    \caption{Results on the fundus segmentation task. Results of the first block are cited from~\cite{chen2021source}. We \textbf{highlight} the best results of each column. ``-'' denotes that the results are not reported and * denotes our implementation.}
    \resizebox{\textwidth}{!}{
      \begin{tabular}{l|ccc|ccc}
      \toprule
      \multirow{2}[4]{*}{Methods} & \multicolumn{3}{c|}{DSC [\%] $\uparrow$} & \multicolumn{3}{c}{ASD [pixel] $\downarrow$} \\
  \cmidrule{2-7}          & Disc  & Cup   & Avg.   & Disc  & Cup   & Avg. \\
      \midrule
      Source & 83.18$\pm$6.46 & 74.51$\pm$16.40  & 78.85 & 24.15$\pm$15.58 & 14.44$\pm$11.27 & 19.30 \\
      BEAL~\cite{wang2019boundary}   & 89.80           & \textbf{81.00}              & 85.40  &       -          &    -             & -  \\
      AdvEnt~\cite{vu2019advent} & 89.70$\pm$3.66  & 77.99$\pm$21.08 & 83.86 & 9.84$\pm$3.86   & 7.57$\pm$4.24   & 8.71 \\
      SRDA~\cite{bateson2020source}   & 89.37$\pm$2.70  & 77.61$\pm$13.58 & 83.49 & 9.91$\pm$2.45   & 10.15$\pm$5.75  & 10.03 \\
      DAE~\cite{karani2021test}    & 89.08$\pm$3.32 & 79.01$\pm$12.82 & 84.05 & 11.63$\pm$6.84  & 10.31$\pm$8.45  & 10.97 \\
      DPL~\cite{chen2021source}    & 90.13$\pm$3.06 & 79.78$\pm$11.05 & 84.96 & 9.43$\pm$3.46   & 9.01$\pm$5.59   & 9.22 \\
      \midrule
      Source * & 88.34$\pm$4.48 & 71.35$\pm$22.75 & 79.85$\pm$12.59 & 10.65$\pm$4.27 & 10.75$\pm$5.34 & 10.70$\pm$4.18 \\
      EMD~\cite{Liu2022}   & 90.50$\pm$3.78 & 73.50$\pm$11.56 & 82.00$\pm$8.76 & 10.52$\pm$4.18 & \textbf{7.12$\pm$4.15} & 8.82$\pm$2.59 \\
      Ours  & \textbf{94.78$\pm$2.65} & 77.79$\pm$12.33 & \textbf{86.28$\pm$7.04} & \textbf{4.41$\pm$2.09} & 8.75$\pm$5.27 & \textbf{6.58$\pm$3.14} \\
      \bottomrule
      \end{tabular}
      }
    \label{tab:results_fundus}
  \end{table}

\begin{table}[t]
  \centering
  \caption{Results on the cardiac dataset and prostate dataset in terms of DSC.}
    \begin{tabular}{l|cccc|c}
    \toprule
    \multirow{2}[4]{*}{Methods} & \multicolumn{4}{c|}{Cardiac}  & \multirow{2}[4]{*}{Prostate} \\
\cmidrule{2-5}          & RV    & Myo   & LV    & Avg.   &  \\
    \midrule
    Source & 40.28$\pm$26.73  & 48.83$\pm$10.84 & 76.45$\pm$10.21 & 55.19$\pm$14.19 & 47.50$\pm$26.21 \\
    EMD \cite{Liu2022}  & 47.59$\pm$28.46 & 53.67$\pm$9.79 & 75.48$\pm$9.58 & 58.91$\pm$13.48 & 52.47$\pm$23.18 \\
    Ours  & \textbf{51.10$\pm$24.67} & \textbf{55.45$\pm$8.88} & \textbf{77.12$\pm$9.01} & \textbf{61.22$\pm$12.41} & \textbf{56.12$\pm$20.72} \\
    \bottomrule
    \end{tabular}
  \label{tab:results2_dice}
\end{table}

\begin{figure}[t]
    \includegraphics[width=\textwidth]{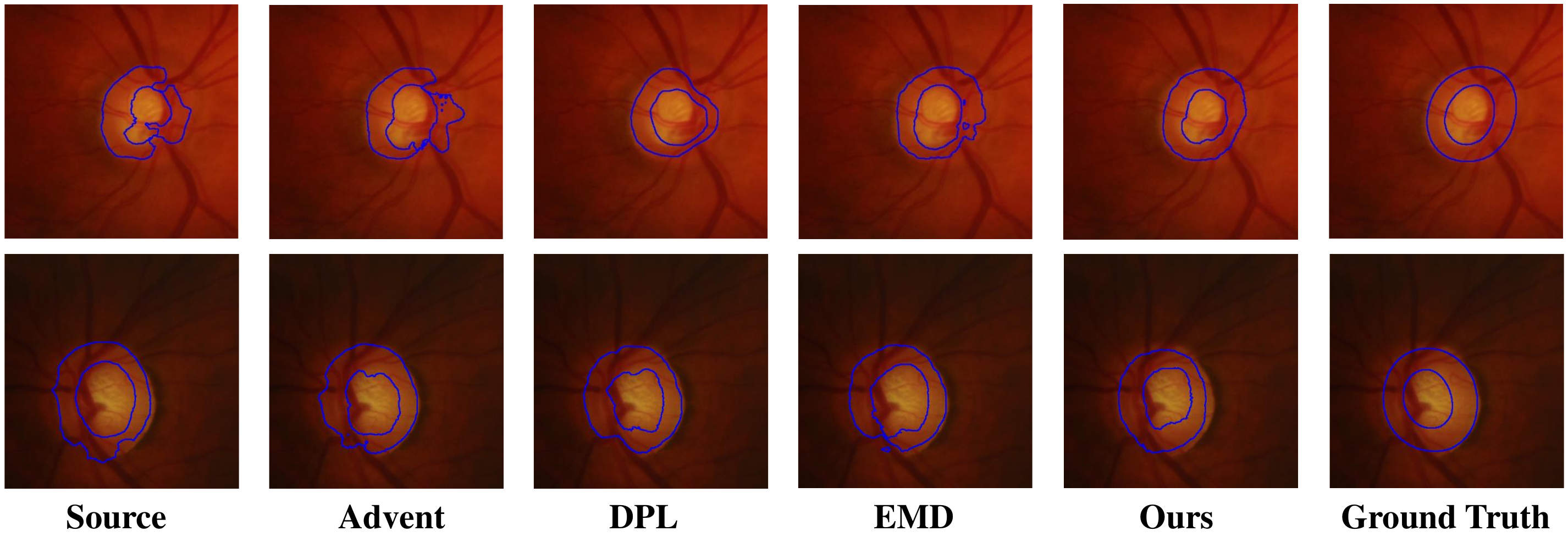}
    \caption{Qualitative results of different methods on RIM-ONE-r3 dataset.}
    \label{fig:visual_results}
\end{figure}

\begin{table}[t]
  \centering
  \caption{Results on the cardiac dataset and prostate dataset in terms of ASD.}
    \begin{tabular}{l|cccc|c}
    \toprule
    \multirow{2}[4]{*}{Methods} & \multicolumn{4}{c|}{Cardiac}  & \multirow{2}[4]{*}{Prostate} \\
\cmidrule{2-5}          & RV    & Myo   & LV    & Avg   &  \\
    \midrule
    Source & 4.50$\pm$3.42 & 4.60$\pm$2.51 & 5.78$\pm$2.02 & 4.96$\pm$1.74 & 9.80$\pm$8.84 \\
    EMD \cite{Liu2022} & 2.12$\pm$1.47 & 4.25$\pm$1.95 & 5.13$\pm$2.78 & 3.83$\pm$1.41 & 8.11$\pm$7.85 \\
    Ours  & \textbf{1.21$\pm$1.05} & \textbf{3.98$\pm$1.56} & \textbf{5.01$\pm$1.95} & \textbf{3.40$\pm$1.32} & \textbf{6.12$\pm$5.42} \\
    \bottomrule
    \end{tabular}
  \label{tab:results2_asd}
\end{table}
\section{Experiments}
\subsection{Dataset}
We evaluate our method on three types of dataset: fundus, cardiac and prostate dataset.

\noindent
\textbf{Fundus}.  We choose the training set of REFUGE challenge \cite{orlando2020refuge} as the source domain and use the public dataset RIM-ONE-r3~\cite{fumero2011rim} as the target domain. Following~\cite{chen2021source,wang2019boundary}, we split the source domain into 320/80 and target domain into 99/60 for training and test, respectively. We resize each image to $512\times 512$ disc region to feed the network during training and test process. 

\noindent
\textbf{Cardiac}.  We use ACDC dataset~\cite{BernardLZCYHCLC18} (200 volumes) as the source domain and  LGE dataset (45 volumes) from Multi-sequence Cardiac MR segmentation Challenge (MSCMR 2019)~\cite{Zhuang19} as the target domain.
For two datasets, we randomly split them into 80\% and 20\% for training and test. We use 2d slices for training and resize all images to $192\times 192$ as the network input.

\noindent
\textbf{Prostate}.  MSD05 \cite{Antonelli2022} (32 volumes) is used as the source domain and Promise12 \cite{litjens2014evaluation} (50 volumes) is used as the target domain. We randomly split 80\%/20\% for training and test, respectively. We use 2d slices for training and resize all images to $224\times 224$ during the training process.
\begin{table}[t]
  \centering
  \caption{Results on \textit{different} networks between source domain and target domain on RIM-ONE-r3 dataset. Source model: DeepLabv3+~\cite{deeplabv3+} with MobileNetV2~\cite{mobilenetv2} backbone (\# param: 5.8 M). Target model: UNet~\cite{unet} (\# param: 1.8 M).}
    \resizebox{\textwidth}{!}{
    \begin{tabular}{l|ccc|ccc}
    \toprule
    \multirow{2}[4]{*}{Methods} & \multicolumn{3}{c|}{Dice [\%] $\uparrow$} & \multicolumn{3}{c}{ASD [pixel] $\downarrow$} \\
\cmidrule{2-7}          & Cup   & Disc  & Avg.   & Cup   & Disc  & Avg. \\
    \midrule
    Stage \uppercase\expandafter{\romannumeral1}  & 46.39$\pm$34.86 & 75.06$\pm$27.98 & 60.73$\pm$29.42 & 14.90$\pm$11.52 & 15.61$\pm$14.45 & 15.26$\pm$8.95 \\
    Stage \uppercase\expandafter{\romannumeral2} w/o aug & 48.45$\pm$33.66 & 76.18$\pm$25.42 & 62.32$\pm$27.4 & 13.78$\pm$10.15 & \textbf{12.65$\pm$11.48} & 13.22$\pm$8.54 \\
    Stage \uppercase\expandafter{\romannumeral2} w/ aug & \textbf{54.22$\pm$33.13} & \textbf{75.89$\pm$28.28} & \textbf{65.05$\pm$28.75} & \textbf{13.27$\pm$9.76} & 12.81$\pm$10.63 & \textbf{13.04$\pm$8.87} \\
    \hline
    EMD~\cite{Liu2022}   & 47.14$\pm$32.56 & 75.48$\pm$26.77 & 61.31$\pm$28.65 & 13.54$\pm$10.88 & 14.75$\pm$12.64 & 14.15$\pm$8.78 \\
    \bottomrule
    \end{tabular}
    }
  \label{tab:results_unet}%
\end{table}%

\begin{table}[t]
  \centering
  \tabcolsep=0.4em
  \caption{\textbf{Ablation Study} on three datasets. Metric: DSC (\%). ``aug'' denotes strong augmentation in Stage  \uppercase\expandafter{\romannumeral2} as introduced in Sec. \ref{sec:two_stage_kd}.}
    \begin{tabular}{cccccc}
    \toprule
    Stage \uppercase\expandafter{\romannumeral1}  &  Stage \uppercase\expandafter{\romannumeral2}  &  w/ aug & Fundus & Cardiac & Prostate \\
    \midrule
          &       &       & 79.85$\pm$12.59 & 55.19$\pm$14.19 & 47.50$\pm$26.21 \\
    \Checkmark     &       &       & 81.59$\pm$9.83 & 58.87$\pm$13.37 & 51.68$\pm$24.56 \\
    \Checkmark     & \Checkmark     &       & 83.61$\pm$8.14 & 59.46$\pm$13.24 & 53.87$\pm$22.99 \\
    \Checkmark     & \Checkmark    & \Checkmark     & \textbf{86.28$\pm$7.04} & \textbf{61.22$\pm$12.41} & \textbf{56.12$\pm$20.72} \\
    \bottomrule
    \end{tabular}%
  \label{tab:ablation}%
\end{table}%

\subsection{Implementation}
Following~\cite{chen2021source}, we choose DeepLabv3+~\cite{deeplabv3+} with MobileNetV2~\cite{mobilenetv2} as the backbone. We use Adam optimizer~\cite{adam} with learning rate as $1e^{-4}$ and set batch size as 8. We train the source model for 200 epochs, and subsequently train the target model for two stages, with each stage consisting of 100 epochs. The weak augmentation only includes Gaussian noise and the strong augmentation includes Gaussian blur, contrast adjustment, brightness, gamma augmentation. For more details of data augmentation, we refer readers to \cite{Isensee2020}. All experiments are implemented using PyTorch on one RTX A6000 GPU.

For evaluation, we use two commonly used metrics in the medical image segmentation task including Dice Score (DSC) for pixel-wise measure 
and the Average Surface Distance (ASD) for measuring the performance at the object boundary. Note that the higher DSC and lower ASD means better performance.

\subsection{Comparison Study}
We mainly compared our method with: \textbf{BEAL}~\cite{wang2019boundary} and \textbf{AdvEnt}~\cite{vu2019advent}, two domain adaptation methods;
\textbf{SRDA}~\cite{bateson2020source}, \textbf{DAE}~\cite{karani2021test} and \textbf{DPL}~\cite{chen2021source}, three source-free domain adaptation methods; \textbf{EMD} \cite{Liu2022}, recent state-of-the-art methods for black-box source free domain adaption.

The quantitative results on the fundus dataset are listed in Table \ref{tab:results_fundus}. From Table \ref{tab:results_fundus}, our method outperforms the baseline (source in Table \ref{tab:results_fundus}) with a significant margin. Specifically, our method outperforms baseline 6.43\% and 4.12 pixels in DSC and ASD. Compared with another black-box method EMD, our method outperforms it by 4.28\% DSC and 2.27 pixels in terms of ASD, respectively. Furthermore, our method defeats all compared methods, even domain adaptation methods (e.g. BEAL~\cite{wang2019boundary} and AdvEnt~\cite{vu2019advent}). Experimental results demonstrate the effectiveness of our approach.

We present qualitative results in Figure \ref{fig:visual_results}, which shows segmentation results of two cases on the fundus dataset. It is observed that our method generates more compact and accurate predictions than compared methods.

Furthermore, we evaluated our method on cardiac and prostate dataset and results are presented in Table \ref{tab:results2_dice} in terms of DSC and Table \ref{tab:results2_asd} in terms of ASD, respectively. We could see that the proposed method still improves baseline stably on the cardiac and prostate dataset. Furthermore, compared with the recent black-box source-free domain adaptation method EMD~\cite{Liu2022}, our method distinctly outperforms EMD 2.3\% and 3.7\% DSC on two datasets, respectively.

Lastly, we evaluate our method in the setting where target users only have limited computation resources. To simulate the situation of low-source target users, we adopted DeepLabv3+~\cite{deeplabv3+} with MobileNetV2~\cite{mobilenetv2} backbone (\# param: 5.8 M) as the source model, and use UNet~\cite{unet} (\# param: 1.8 M) as the target model. We keep the other training details the same as in previous experiments. The quantitative results of fundus dataset are listed in Table \ref{tab:results_unet}. It is seen that our method still outperforms EMD~\cite{Liu2022} by a significant margin. More importantly, this experiment demonstrates the flexibility of black-box source-free domain adaptation when target users have limited computation resources. \textit{Without any groundtruth on the target domain} , the small target model (e.g. UNet with 1.8 M parameters) will achieve promising results with 65.05\% DSC and 13.04 pixel ASD only using outputs from the (bigger) source model.

\subsection{Ablation Study}
In this section, we conduct ablation studies on three datasets. The quantitative results are listed in Table \ref{tab:ablation}. From Table \ref{tab:ablation}, it is noticed that the target model only trained with Stage \uppercase\expandafter{\romannumeral1} improves baseline with a significant margin. Specifically, Stage \uppercase\expandafter{\romannumeral1} brings 1.7\%, 3.7\%, and 4.2\% DSC gains compared with baseline. Furthermore, the performance of target models could be improved stably in Stage \uppercase\expandafter{\romannumeral2}. Finally, with the help of strong data augmentation, we achieved the best performance on three datasets. This demonstrates the importance of strong data augmentation.

\section{Conclusion}
In this paper, we address the challenging yet practical problem: black-box source-free domain adaptation. We propose a new method to tackle this problem. The key idea is that we aim to transfer knowledge from the black-box source model to the target model with two-stage knowledge distillation. In Stage \uppercase\expandafter{\romannumeral1}, we transfer knowledge from the black-box source model to the target model. In Stage \uppercase\expandafter{\romannumeral2}, we regard the target model as the new teacher to guide the learning of augmented images. Thanks to the two-stage knowledge distillation, our model could achieves remarkable performance on the target domain without any ground truth. Finally, we conducted extensive experiments on three medical image segmentation tasks, and the results demonstrate the effectiveness of the proposed method.
%\clearpage
\bibliographystyle{splncs04.bst}
\bibliography{reference}
\end{document}